# Application of sequential processing of computer vision methods for solving the problem of detecting the edges of a honeycomb block

**M V Kubrikov[1], I A Paulin[1], M V Saramud[1, 2] and A S Kubrikova[1]**

[1]Reshetnev Siberian State University of Science and Technology, 31, Krasnoyarsky Rabochy Av., Krasnoyarsk, 660037, Russia
[2]Siberian Federal University, 79, Svobodny pr., Krasnoyarsk, 660041, Russia

E-mail: msaramud@gmail.com

**Abstract.** The article describes the application of the Hough transform to a honeycomb block image. The problem of cutting a mold from a honeycomb block is described. A number of image transformations are considered to increase the efficiency of the Hough algorithm. A method for obtaining a binary image using a simple threshold, a method for obtaining a binary image using Otsu binarization, and the Canny Edge Detection algorithm are considered. The method of binary skeleton (skeletonization) is considered, in which the skeleton is obtained using 2 main morphological operations: Dilation and Erosion. As a result of a number of experiments, the optimal sequence of processing the original image was revealed, which allows obtaining the coordinates of the maximum number of faces. This result allows one to choose the optimal places for cutting a honeycomb block, which will improve the quality of the resulting shapes.

## 1. Introduction
Manufacturing honeycomb blocks of complex shape is a laborious technological task [1]. The complexity of this task lies in the formation of the trajectory of the cutting tool and restrictions on the place and angle of the cut. The vertices of the polygon should not fall under the cut, since this can lead to a poor-quality cut and deterioration of the strength characteristics of the honeycomb block [2]. For optimal cutting of the honeycomb block, it is necessary to cut in the right place and at the right angle. The calculation of the optimal cutting locations is a separate scientific problem. However, to solve this problem, we need to have a digital image of the canvas of the honeycomb block. That is, for calculations, we need the coordinates of the nodal points and the location of the faces of the honeycomb block. This data cannot be predicted because the structure of the real honeycomb block is irregular. Honeycomb blocks are made by gluing and stretching aluminum strips. As a result of the stretching of the glued tapes, the individual cells are stretched unevenly, thus the geometry of the honeycomb block is not geometrically uniform. Thus, the problem arises of obtaining the real coordinates of the nodal points and faces for the possibility of further calculations. One of the ways to solve this problem is to use machine vision methods.

## 2. The problem of processing low quality images
Due to the technological features of the location of the honeycomb block on the working surface of the machine [3], it is not always possible to create an optimal background for photography under the honeycomb block. There is also parasitic illumination of the subject by general light in the room. Due to these reasons, shooting may result in poor quality images. If the image quality is not good enough as in Fig. 1, problems arise in detecting the honeycomb and hence determining the path for the cutting blade. Due to optical distortions closer to the edges of the picture, the walls of the honeycomb change the perspective angle, which introduces significant distortions into the real geometry of the honeycomb block and, as a consequence, the quality of determining the honeycomb structure of the block.

Using the OpenCV library [4] and the Python language, we will conduct a study and identify the most effective methods for solving this problem.

To construct cells, we use the Hough transform [5, 6]. The Hough transform is a technique for detecting straight and curved lines in grayscale or color images. The method allows you to specify the parameters of a family of curves and provides a search on the image for a set of curves of a given family. We will consider its application to find straight line segments in an image.

If we apply this method to the original image, we get the following result, see figure 1 on the right.

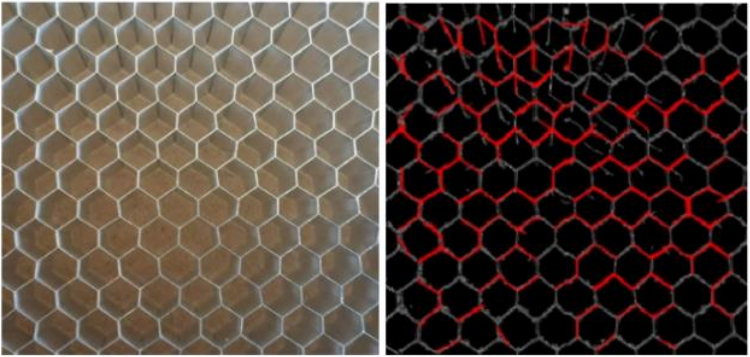

**Figure 1.** Original image of honeycomb block (left) and Hough transform (right).

The reason why we get an image different from the original is that before applying the Hough transform, the original image undergoes a series of transformations, namely, converting the image into a single-channel image in grayscale and applying the Canny method [7], which outlines the perimeter of objects located on the image. As a result, we have double lines throughout the honeycomb structure.

We can observe that not all the faces shown in the image are detected, and also, due to noise in the photo, false lines are detected.

To solve this problem, we will apply the method of obtaining a binary image using a simple threshold (static binarization).

This method consists in statically setting the threshold value, up to which all pixel brightness values become 0, and after the threshold they take a value of 255.

After using this method, we get the result shown in figure 2 on the left. Let's apply the Hough transformation to this image (figure 2 on the right).

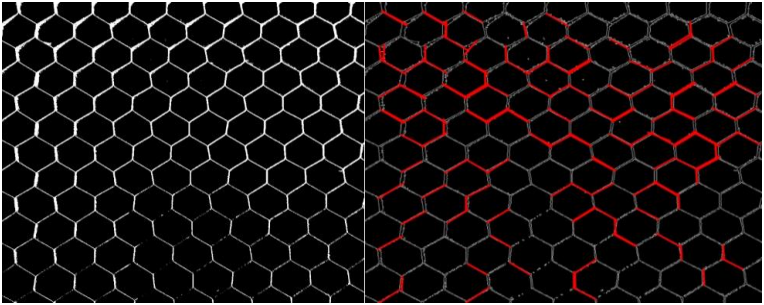

**Figure 2.** Binary Threshold Image (left) and Hough Transform (right).

As a result, one can see that this method helped us get rid of false lines, but not all faces are detected.

Next, we will apply the method of obtaining a binary image using OTSU binarization. Otsu's method avoids the need to select a threshold value and determines it automatically. An image with only two different image values (bimodal image), where the histogram will consist of only two peaks. The optimal threshold is in the middle of these two values. Likewise, OTSU's method [8] determines the optimal global threshold value from the image histogram.

To do this, use the cv.threshold () function, where cv.THRESH_OTSU is passed as an optional flag. The threshold value can be freely selected. The algorithm then finds the optimal threshold value, which is returned as the first output.

We apply this method to our image and the Hough transformation, we get the result shown in figure 3.

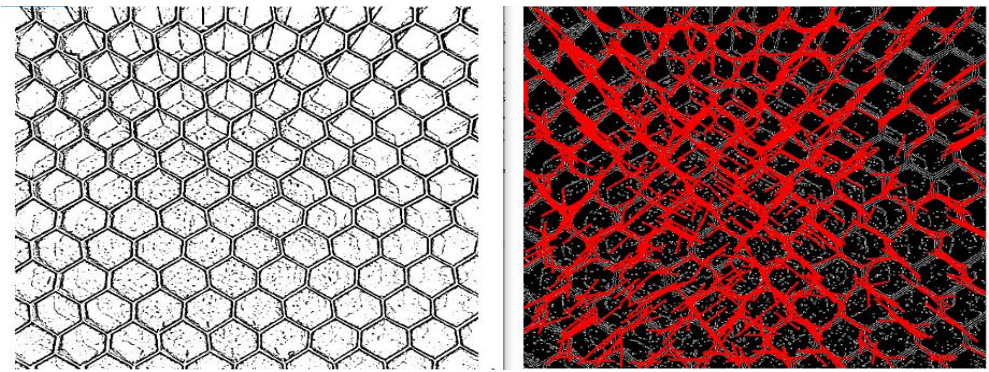

**Figure 3.** OTSU binarization (left) and applying the Hough transform (right).

As a result, it becomes obvious that this method gives a good visual picture but is not at all suitable for detecting straight lines for our task due to the many false positives.

Next, we will try the method of obtaining an image using the Canny algorithm [9]. Canny Edge Detection is a popular edge detection algorithm. It was designed by John F. Canny in 1986.

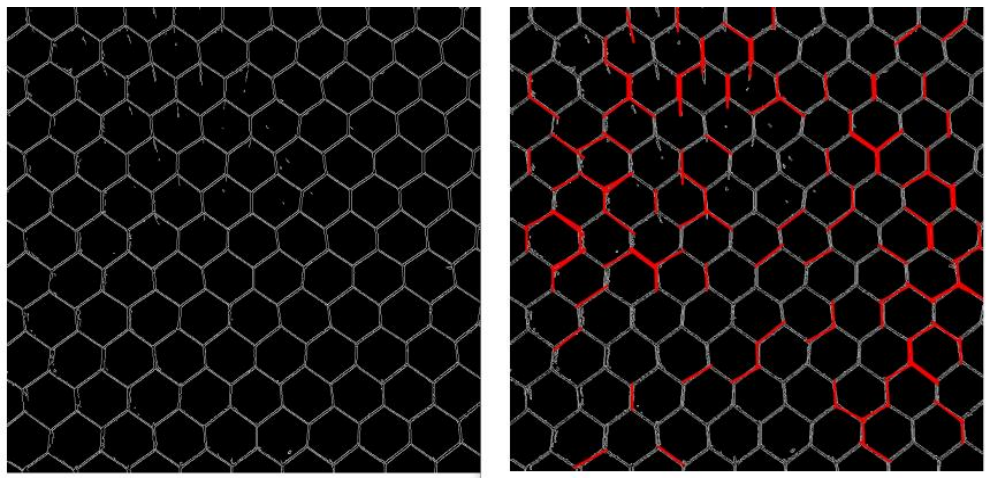

**Figure 4.** Image after applying Canny (left) and applying Hough transform to it (right).

As one can see in figure 4, detection is even worse than with binarization with a simple threshold, and false lines also appear.

To eliminate noise in the image, we use the erosion method. The main idea of the method is similar to soil erosion; it blurs the boundaries of the foreground object. From the entire array of pixels in the image, we select the "core" - an area of 3x3 or 5x5 pixels around the pixel in question. The algorithm of the method consists in sequential sliding of the kernel over the image. The pixel value in the original image will be considered equal to 1 only if the values of all pixels under the kernel are equal to 1, otherwise it will be blurred (zeroed). All pixels near the border will be discarded, depending on the size of the kernel. Thus, the thickness or size of the foreground object is reduced, or simply the white area in the image is reduced. This is useful for removing small white noise.

Usually in cases such as noise removal, erosion is followed by expansion. Because erosion removes white noise, but it also compresses our object. Let's try to apply the extension (Dilation). In this method, the pixel value is "1" if at least one-pixel value under the kernel is "1". This algorithm of work increases the white area in the image or increases the size of the foreground object. Using this method, we expand the objects in the image. Since small objects have been completely removed, only the area of our object will increase. It also helps to connect the broken parts of the object.

Since, from the previous methods, we had the best result with static binarization, we apply this method to it (figure 5 and figure 6).

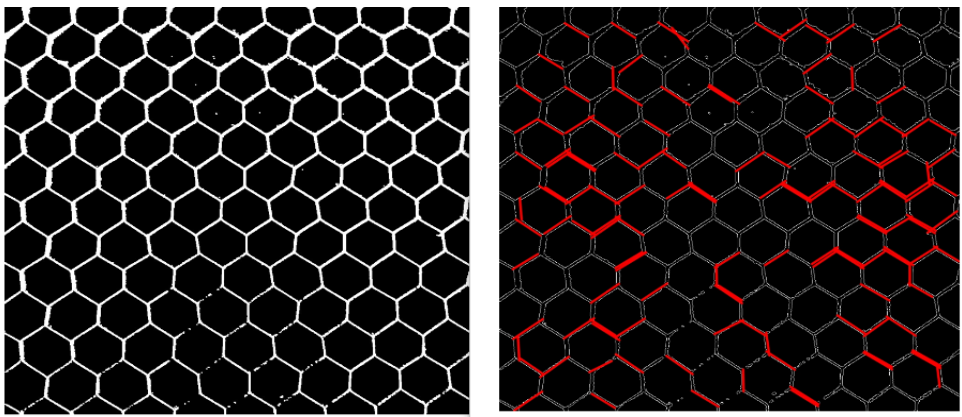

**Figure 5.** Image of static binarization after one expansion step (left) and Hough transformation (right).

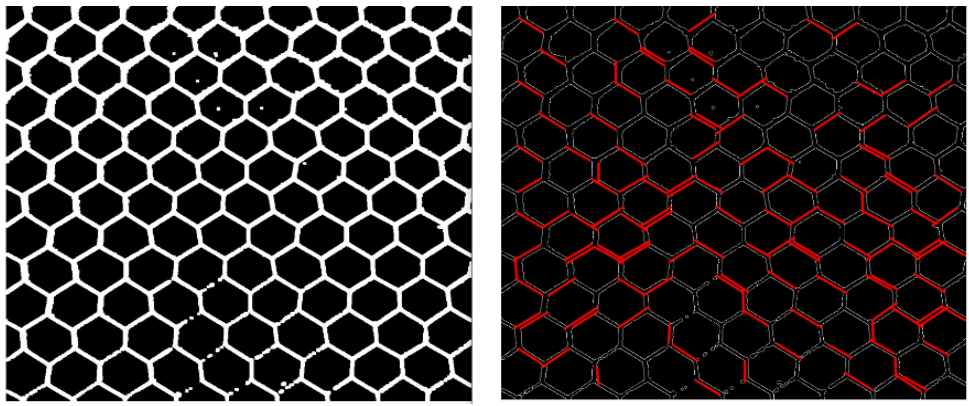

**Figure 6.** Image of static binarization after two expansion steps (left) and Hough transform (right).

One can see that there was no significant increase in the number of detected lines, but duplicated ones were obtained.

Next, let us try to get an image using a binary skeleton (skeletonization) [10].

The skeleton is obtained using 2 main morphological operations: Dilation and Erosion.

As an example, take an image obtained using Otsu binarization [11] and carry out erosion over it, see figure 7.

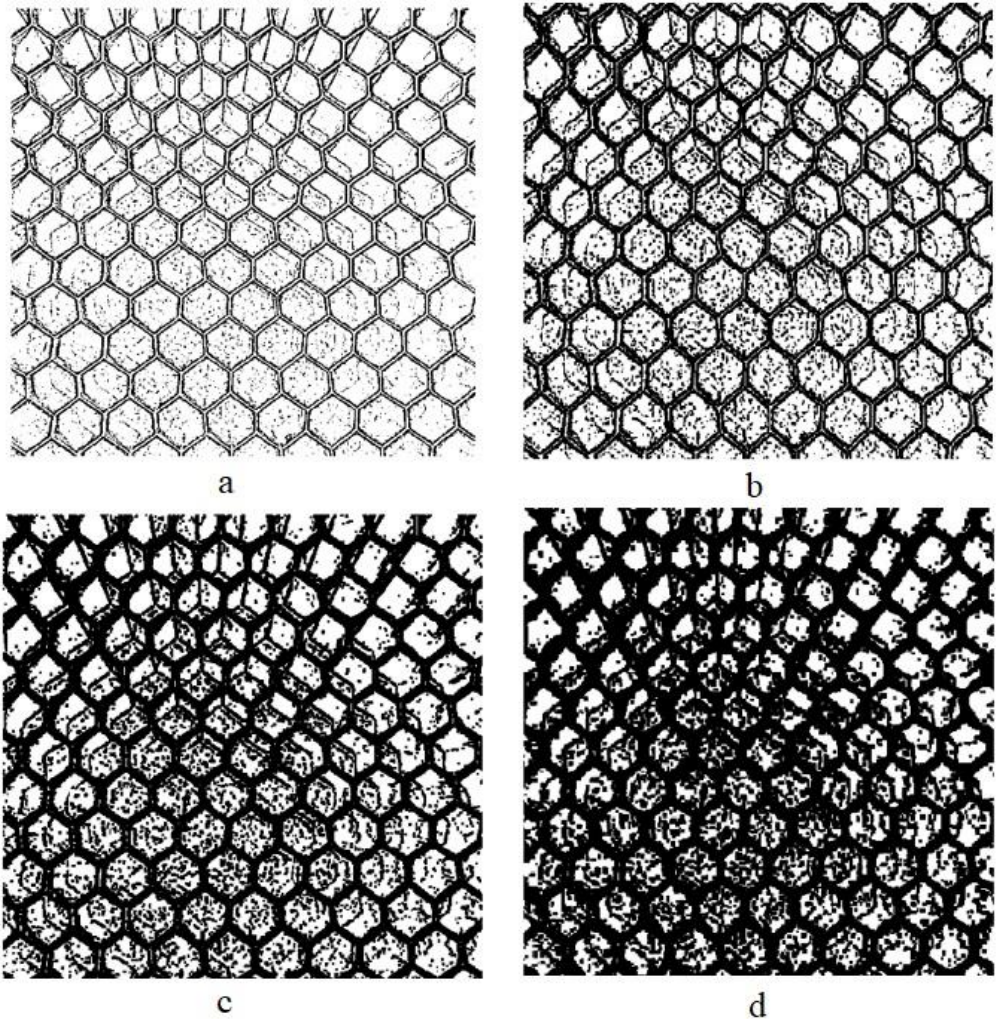

**Figure 7.** Applying erosion to an image after OTSU binarization, a) to an image after OTSU binarization, and b) an image after one step of erosion c)) an image after two steps of erosion d)) an image after three steps of erosion.

Let us describe the algorithm for obtaining an image skeleton:

1. We start with an empty skeleton;
2. We remove noise;
3. Subtract from the original image the one that was obtained in the second paragraph;
4. Removing the original image and obtaining a skeleton by combining the current skeleton with the image obtained in step 3;
5. Repeat steps 2-4 until the original image is deleted.

Let us build a skeleton for the original image and apply the Hough transformation to it (figure 8).

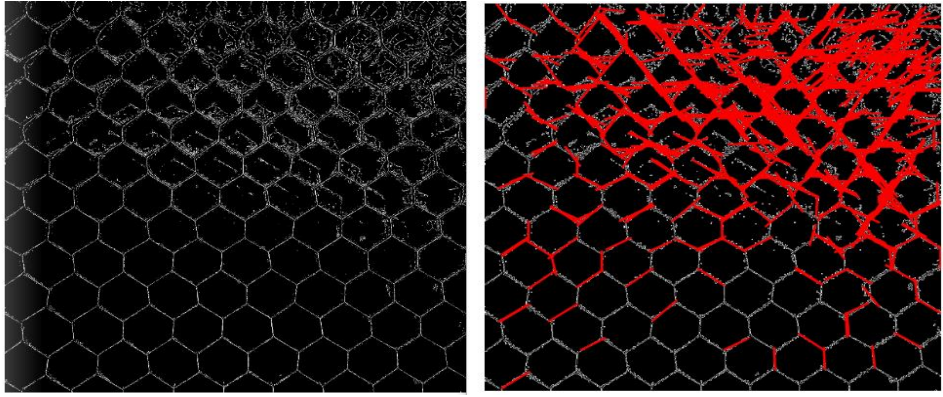

**Figure 8.** Get the skeleton (left) and apply the Hough transform (right).

As one can see, this method applied to the original image does not bring the desired result. Let us build a skeleton for an image with static binarization (figure 9).

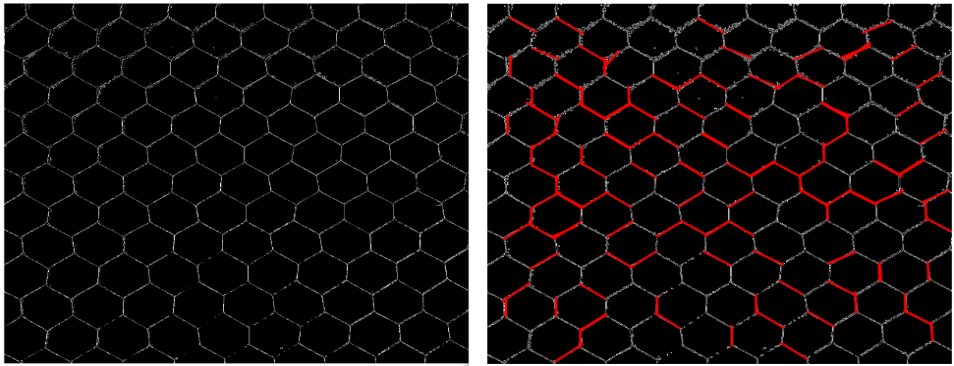

**Figure 9.** Get the skeleton (left) and apply the Hough transform (right).

As we can see so far this is the best result, now we will apply the following algorithm:

- Let us make an extension with a 5x5 kernel for the image obtained using static binarization.
- Let us apply erosion with a 3x3 core. Let's build a skeleton. The result can be seen in figure 10 and figure 11.

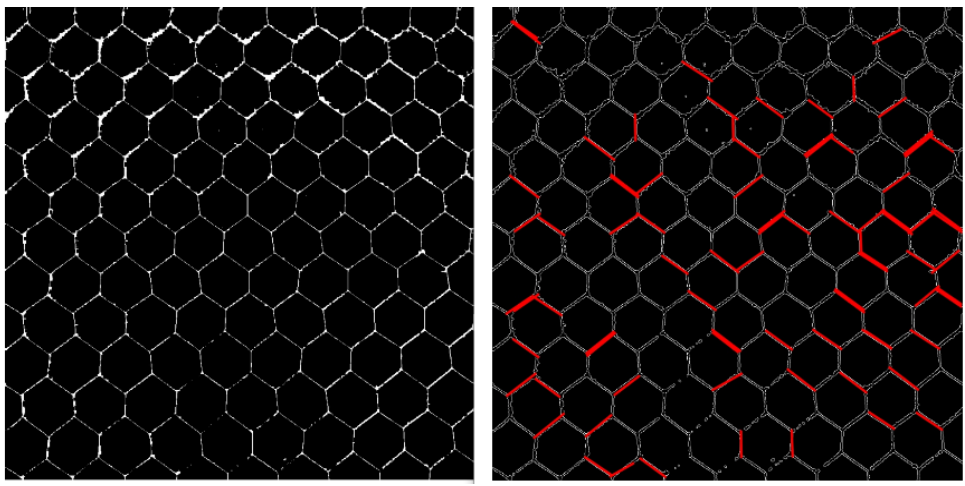

**Figure 10.** Obtaining an image at 1 step of erosion.

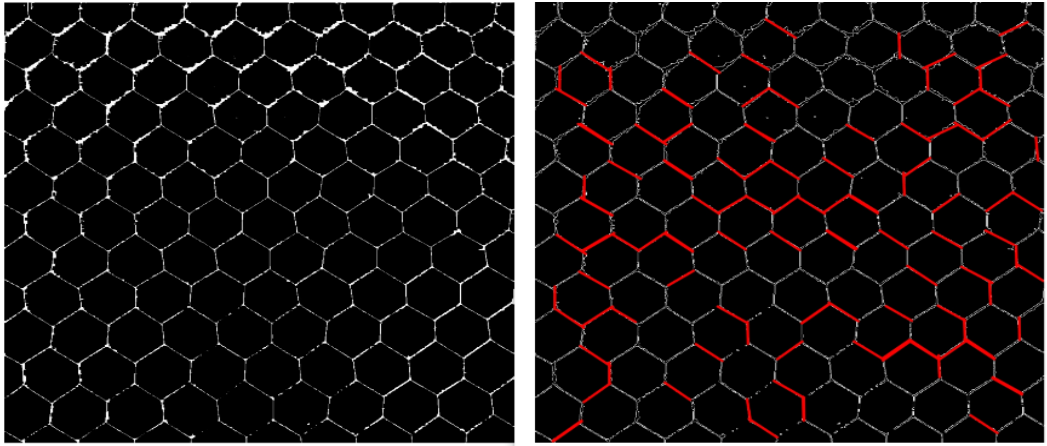

**Figure 11.** Acquisition of an image at 2 steps of erosion.

After the second step, the detection is better, but some of the lines found in the first step were not found in the second. Further application of erosion for this image is not advisable because most of the contours are overwritten.

To obtain a larger number of coordinates of the lines found, summarize the result in figures 9, 10 and 11. The result is shown in figure 12.

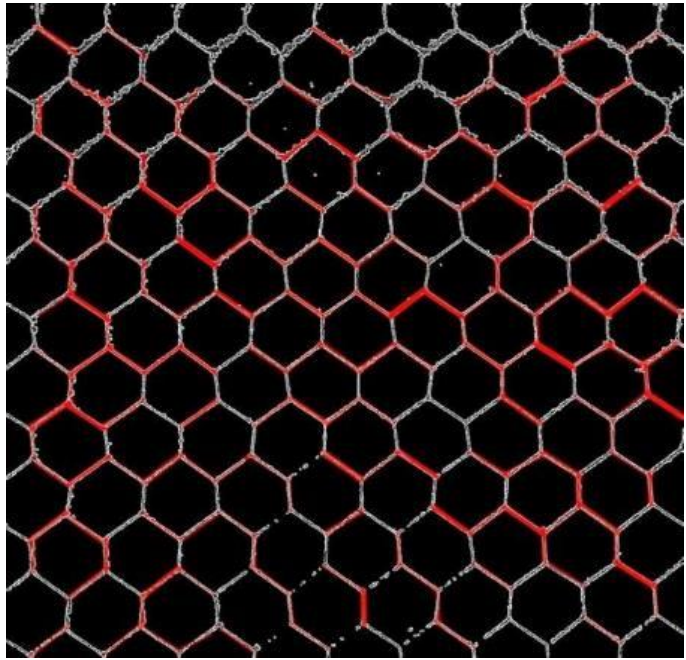

**Figure 12.** Sum of figures. 9, 10 and 11.

After analyzing the results after each operation separately and in combination, an algorithm was obtained, which is a combination of methods. This algorithm produces the best possible result.

Let us present the algorithm in the form of a block diagram in figure 13.

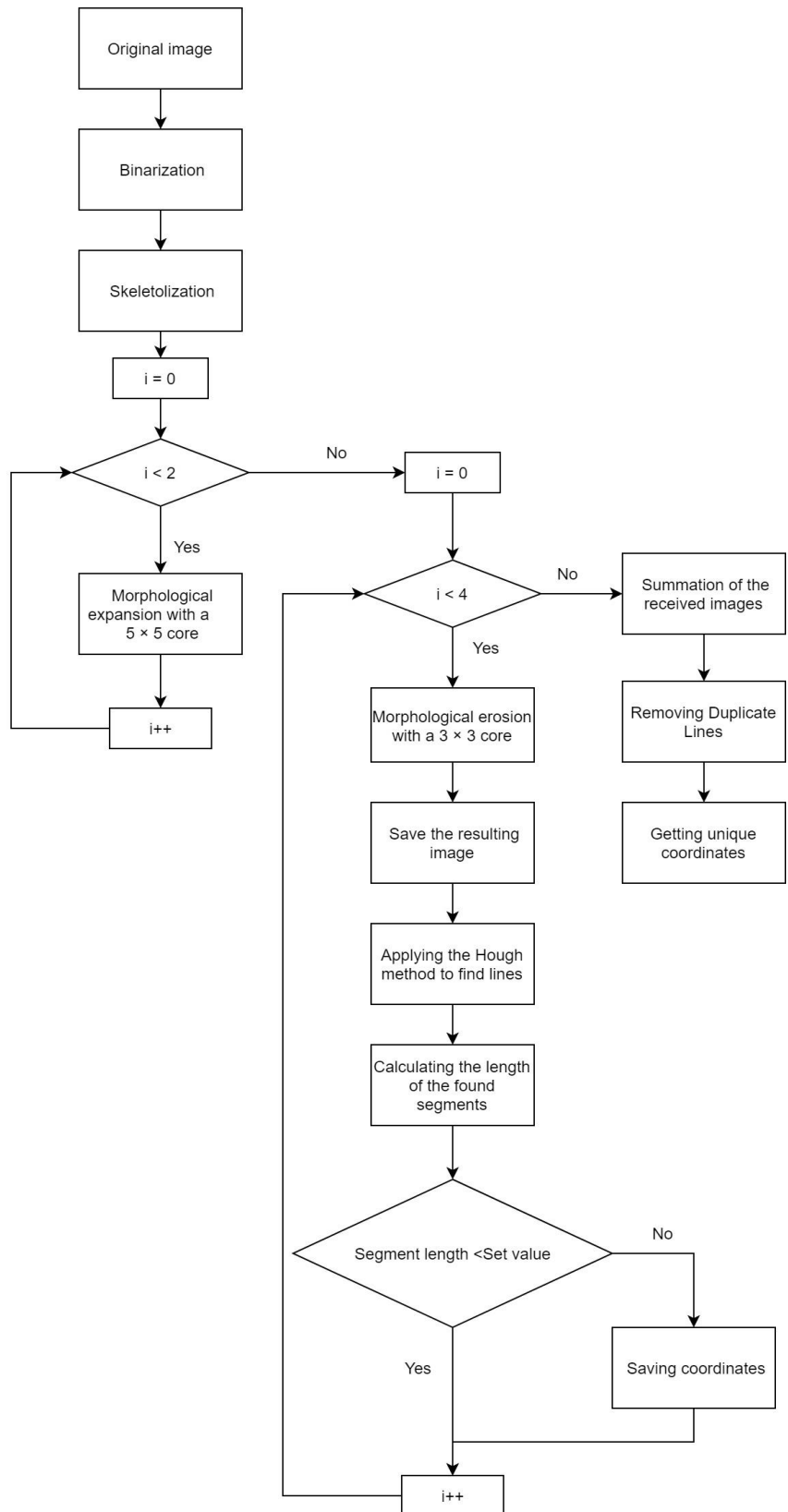

**Figure13.** Block diagram of the algorithm.

This algorithm is easy to use because it can be implemented using the OpenCV library. This library is available for C / C ++, python, java. Among the shortcomings of the obtained algorithm, it is worth noting the need to calibrate the number of loop iterations during the first starts or when changing equipment to obtain images.

**4. Conclusion**
The use of a sequence of actions and various combinations of machine vision methods makes it possible to obtain coordinates on average about 97% of the edges of the honeycomb structure. Knowing the coordinates of the faces, it is possible to determine the path for the cutting tool and exclude nodal sections from the cutting zone of the honeycomb. The result obtained as a result of the study will allow you to choose the optimal places and the angle of the knife for cutting the honeycomb block. The optimal knife positioning scheme will improve the quality of the resulting shapes.

**Acknowledgements**
This work was supported by the Ministry of Science and Higher Education of the Russian Federation (State Contract No. FEFE-2020-0017).